\documentclass[10pt,twocolumn,letterpaper]{article}
\usepackage[pagenumbers]{iccv}
\usepackage{multirow}
\usepackage{placeins}
\usepackage{setspace}

\definecolor{iccvblue}{rgb}{0.21,0.49,0.74}
\usepackage[pagebackref,breaklinks,colorlinks,allcolors=iccvblue]{hyperref}
\usepackage{colortbl}

\usepackage[activate=true,
            protrusion=true, 
            expansion=true, 
            tracking=false,
            final]{microtype}
\microtypecontext{spacing=nonfrench}
\usepackage{graphicx}
\usepackage{afterpage}

\title{DANTE-AD: Dual-Vision Attention Network for Long-Term Audio Description}

\author{
\textbf{Adrienne Deganutti, \quad Simon Hadfield, \quad Andrew Gilbert} \\
University of Surrey, UK \\
{\tt\small \{a.deganutti, s.hadfield, a.gilbert\}@surrey.ac.uk}
}

\begin{document}
\maketitle
\begin{abstract}
Audio Description is a narrated commentary designed to aid vision-impaired audiences in perceiving key visual elements in a video. While short-form video understanding has advanced rapidly, a solution for maintaining coherent long-term visual storytelling remains unresolved. Existing methods rely solely on frame-level embeddings, effectively describing object-based content but lacking contextual information across scenes. We introduce DANTE-AD, an enhanced video description model leveraging a dual-vision Transformer-based architecture to address this gap. DANTE-AD sequentially fuses both frame and scene level embeddings to improve long-term contextual understanding. We propose a novel, state-of-the-art method for sequential cross-attention to achieve contextual grounding for fine-grained audio description generation. Evaluated on a broad range of key scenes from well-known movie clips, DANTE-AD outperforms existing methods across traditional NLP metrics and LLM-based evaluations. 
\end{abstract}

\section{Introduction}
\label{sec:intro}
The booming television, streaming, and digital media industries provide entertainment to billions of people worldwide. However, this market is fundamentally designed for those who can see. An estimated 338 million people live with moderate to total blindness \cite{bourne2021trends}, making video-based media inherently inaccessible to them. Beyond the inability to see the screen, these audiences miss out on crucial storytelling elements, such as cinematography, lighting, colour schemes, and visual symbolism—techniques that convey emotions, themes, and narrative depth. Without access to these visual cues, their experience is significantly diminished.

\begin{figure}[t]
\vspace{-0.8cm}
  \centering
  \includegraphics[width=\linewidth]{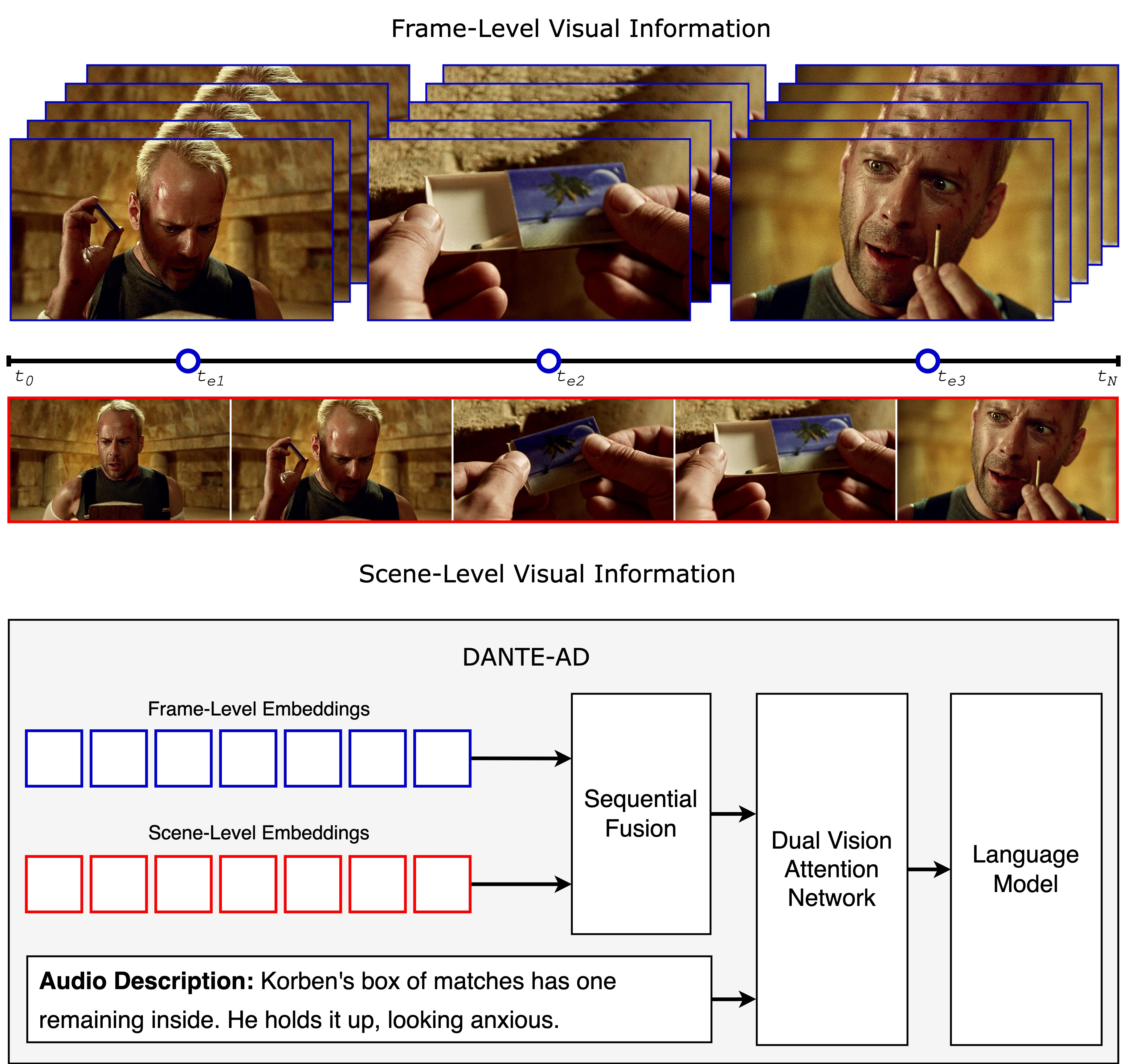}
  \caption{Our DANTE-AD method extracts frame- and scene-level visual information fused via a sequential cross-attention module for both frame and scene context-aware AD over extended video sequences.}
  \label{fig:intro}
\end{figure}

One solution is Audio Description (AD), a critical accessibility service for visually impaired audiences, providing spoken narration that describes on-screen actions, characters, locations, costumes, body language, and facial expressions \cite{ATVoD}. Unlike simple video captioning, which reports only the main actions or events in a video, AD enhances meaning and immersion by incorporating visual context and emotion into the narrative. The growing societal support for accessibility has increased legal requirements for AD across television, film, and streaming platforms. However, manual AD creation remains resource-intensive, requiring skilled professionals with domain expertise and significant time investment. As a result, many videos either lack AD entirely or offer only limited coverage. Automating AD generation is, therefore, a crucial challenge.

\smallskip
Automating AD generation relies on two fundamental capabilities: (i) Recognising and describing objects, actions, and events within a video and (ii) Producing coherent and contextually relevant narrative descriptions. Video captioning techniques primarily focus on the first capability for short video clips, where they effectively capture simple actions \cite{lin2022swinbert}. However, they fall short in fulfilling the second capability, lacking the nuanced detail necessary for coherent and contextually rich descriptions. Furthermore, video captioning becomes inadequate as video duration increases, such as in films and television programmes. It struggles to handle extended temporal sequences due to the added complexity of multiple concurrent events. Therefore, a significant gap persists in generating detailed and context-aware descriptions that capture a broader narrative across long-form video content.

\smallskip
To address these limitations, we propose \textbf{DANTE-AD} (\textbf{D}ual-Vision \textbf{A}ttention \textbf{N}etwork for Long-\textbf{Te}rm \textbf{A}udio \textbf{D}escription)—a video-based representation learning framework that integrates multimodal visual embeddings to improve long-term context awareness for AD generation.

\smallskip
Our motivation is inspired by the narrative principle of Chekhov's Gun \cite{debreczeny1984chekhov}, which states that every element in a story should serve a purpose. In video storytelling, visually represented elements are essential to the plot and character development, meaning that AD must prioritise relevant visual details to ensure narrative comprehension as the story unfolds.

\smallskip
In line with this principle, our model is designed to better identify and retain key visual elements, ensuring that the context is maintained as the narrative unfolds. Our dual-vision approach uniquely incorporates frame and scene level embeddings, providing complementary representations with distinct spatial and temporal modelling strengths. Our \textbf{frame-level} embeddings capture rich semantic content, including objects, scenes, and contextual relationships. These embeddings operate at a high dimensionality with query-based attention and primarily focus on spatial representations. In contrast, our \textbf{scene-level} embeddings capture dynamic information across a more extended video sequence, such as the temporal evolution of actions, object trajectories, and inter-frame dependencies, resulting in a global temporal representation. To our knowledge, our method is the first to employ multi-type visual embeddings for AD generation. Our unique approach integrates these embeddings through sequential fusion, which ensures that the learned context from the scene level information is retained when querying the frame level embeddings. Our method enhances the model's ability to generate detailed and context-aware descriptions over extended video sequences.

\smallskip
\noindent In summary, our key contributions are the following. 
\begin{enumerate}
    \item Leveraging scene- and frame-level visual information to provide complementary representations, each optimized for distinct spatial and temporal modelling strengths.
    \item Incorporation of a Dual-Visual Attention Network within the AD pipeline. Using a sequential fusion technique, our model fuses frame- and scene-level embeddings, enabling the dual-vision attention network to retain global contextual information while querying frame-level representations. This approach yields a more comprehensive video content representation and improves alignment between visual information and generated descriptions.
    \item Extensive evaluation results on real long-term film clips, using realistic real-world metrics.
\end{enumerate}

\begin{figure*}[t]
  \centering
  \includegraphics[width=\linewidth]{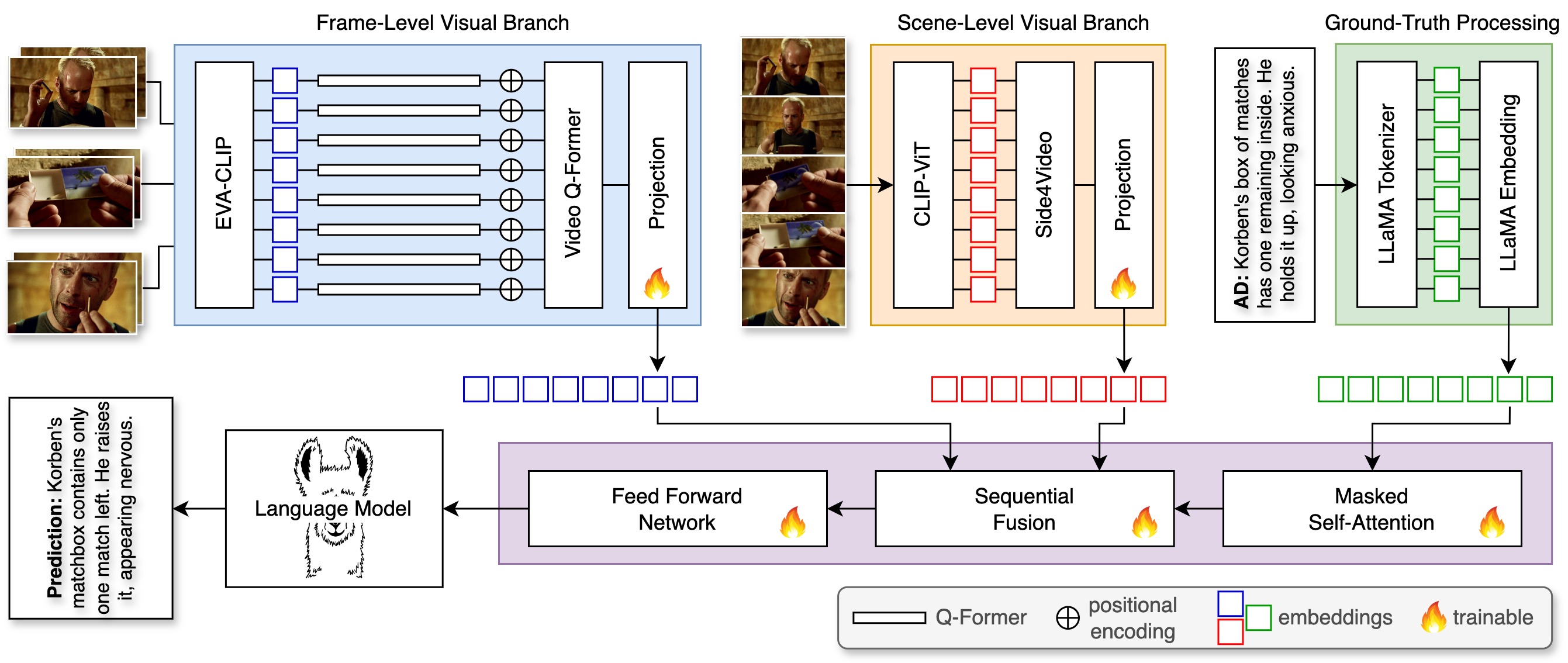}
   \caption{Overview of our audio description generation pipeline. The system features two primary branches: a frame-level visual branch (blue) and a scene-level visual branch (red). Ground-truth references are embedded and processed auto-regressively using a causal attention mask. Sequential fusion integrates the visual embeddings within the Dual-Vision Attention Network (purple). The fused representation is fed to our LLaMA language model and decoded into a natural language AD prediction.}
   \label{fig:model_overview}
\end{figure*}

\section{Related Works}
\label{sec:related_works}
Lying at the intersection of Natural Language Processing (NLP) and computer vision, video captioning generates textual descriptions of video content by modelling the relationship between visual and linguistic information. Early approaches relied on sequence-to-sequence architectures with attention mechanisms \cite{venugopalan2015sequence, yu2016video} to generate temporally coherent descriptions. More recent advancements incorporate transformer-based models \cite{sun2019videobert, lin2022swinbert} and reinforcement learning techniques \cite{pasunuru2017reinforced} to improve fluency and relevance. Additionally, retrieval-augmented approaches \cite{miech2020end, xu2024retrieval} leverage large-scale video-text datasets to enhance contextual accuracy. Despite these improvements, conventional video captioning methods often struggle with generating descriptions that capture the depth of narrative structure and long-term dependencies \cite{han2023autoad1}, which are crucial for audio description (AD) tasks. This limitation motivates the exploration of techniques such as visual storytelling, multimodal integration, and long-term modelling, as discussed in the following sections.

\subsection{Visual Storytelling}

Various machine learning fields rely on visual storytelling to convey complex spatial, narrative, and emotional information, allowing models to interpret and generate visual content contextually and meaningfully. This includes areas such as Vision-Language Navigation (VLN) within 3D environments \cite{wang2019reinforced}, video game generation \cite{liu2021deep}, and persuasiveness analysis within social media \cite{shin2020enhancing}. Specifically, AD leverages visual storytelling to generate an engaging and logically structured explanation of a visual story, one of the main qualities that differentiates AD from traditional video captioning tasks. AD requires cohesion across frames rather than isolated descriptions. It also relies on engaging narration and a logically structured explanation of the visual content. Studies such as \cite{fang2024distinctad} and \cite{you2024towards} have explored removing redundant descriptions across similar frames to enhance narrative flow. While \cite{nagrani2017benedict, han2023autoad2, han2024autoad3} demonstrated how identifying characters within the video improves the perceptual understanding of the story. This clarity enhances the narrative flow and ensures that key interactions and dialogue are clearly attributed. In contrast, our work focuses on a more nuanced understanding of the scene, improving narration and context comprehension. By incorporating dual visual embeddings, our method provides a more detailed portrayal of the visual story.

\subsection{Multimodal Video Captioning}

Video captioning tasks have commonly worked on incorporating additional input modalities to achieve better results, such as audio \cite{shen2023fine, iashin2020multi}, subtitles \cite{lei2020tvr}, dynamic motion \cite{chen2020learning}, external memory banks \cite{kim2024you}, and emotion vocabulary \cite{song2024emotional} among others. These multimodal techniques enable richer descriptions by leveraging diverse input sources. For example, \cite{lei2020tvr} demonstrates moment localisation within a video by leveraging temporally aligned subtitles. Multimodal approaches also improve video-text alignment. For instance, \cite{zhu2015aligning, tapaswi2015book2movie} align movies with their corresponding books, enhancing narrative comprehension. Our method introduces a novel approach to multimodal AD generation by integrating a second visual modality focusing on context, alongside the frame-based visual modality and text-form ground-truth references. This aims to improve both scene understanding and narrative depth.

\subsection{Long-Term Modelling}

Video captioning extends image captioning \cite{you2016image} by incorporating temporal dynamics. Techniques such as sliding window approaches \cite{chen2025sharegpt4video} help capture motion across neighbouring frames. However, most video captioning methods \cite{lin2022swinbert, shen2023fine, you2016image, xu2024retrieval} rely on short-form video datasets \cite{shen2023fine, zhou2018towards, wang2019vatex, xu2016msr}, which results in captions lacking global coherence or long-term narrative structure. Unlike video captioning, generating coherent audio descriptions depends on the ability to capture long-term dependencies, particularly to avoid repetition of information. Recent techniques aim to retain long-term memory efficiently, with methods such as autoregressive decoding \cite{piergiovanni2024whats}, memory clustering \cite{zhou2024streaming}, and recursive video captioning \cite{islam2024video}, which generates captions at multiple hierarchical time scales for more coherent long-form descriptions. Our method addresses these challenges by enhancing scene understanding through dual visual embeddings, allowing for improved long-term dependency modelling as video durations increase.

\section{Model Architecture}
Our model architecture is shown in \cref{fig:model_overview}. It consists of two parallel feature extraction branches, one for extracting and processing frame-level embeddings and another for global scene-level representations. A dual-vision Transformer network sequentially integrates the individual spatial frame details with the long-term temporal contextual scene information to fuse the two sets of embeddings. A frozen Large Language Model decodes the fused information into natural language by decoding the fused logits. 

\subsection{Frame-Level Visual Branch}
\label{subsec:frame-level-visual-branch}
We consider a film sequence consisting of \(N\) frames: 
\begin{equation}
    M = \{I_1, I_2, \dots, I_N\}
    \label{eq:1}
\end{equation}
where each frame \(I_i\) is a 3-channel RGB image of spatial resolution \(H \times W\). To reduce computational cost while retaining temporal information, we uniformly subsample the video to \(T\) frames:
\begin{equation}
    M_T = \{I_{t_1}, I_{t_2}, \dots, I_{t_T}\}, \text{where } t_k = \left\lceil \frac{kN}{T} \right\rceil
    \label{eq:2}
\end{equation}
Given a batch size \(B\), our input video representation \(X\) is such that \({X} \in \mathbb{R}^{B \times 3 \times T \times H \times W}\).

\medskip
\noindent \textbf{Visual Encoder} We extract frame-level dense visual embeddings for each frame independently using a pre trained EVA-CLIP (ViT-G/14) from the pre-trained BLIP-2 model \cite{li2023blip}. The vision encoder outputs per-frame embeddings:
\begin{equation}
    L_{\text{frames}} = f_{\text{vision}}(X) \in \mathbb{R}^{(B \times T) \times D_v}
    \label{eq:3}
\end{equation}
where \(D_v\) is the hidden dimension of the vision model. To create spatial awareness across the embeddings, the pre-trained Q-Former from BLIP-2 \cite{li2023blip} performs cross attention between each frame embeddings and a collection of learnable query tokens, \(Q_F\) to extract additional frame-level spatial information:
\begin{equation}
    Q^{(t)}_{\text{frames}} = \text{Q-Former}(Q_F, L^{(t)}_{\text{frames}}) \in \mathbb{R}^{N_q \times D_q}
    \label{eq:4}
\end{equation}
where \(N_q\) is the number of query tokens and \(D_q\) is their hidden size. To encode temporal information across frames, we add positional embeddings to the learned queries \(E^{(t)}\) that correspond to the relative frame positions, thereby capturing the sequential structure of the video data.
\begin{equation}
    \tilde{Q}^{(t)}_{\text{frames}} = Q^{(t)}_{\text{frames}} + E^{(t)}
\end{equation}

\medskip
\noindent \textbf{Frame Video Q-Former.} The position-encoded frame embeddings are flattened to obtain a global video representation such that \(\tilde{Q}^{(t)} \in \mathbb{R}^{B \times (T \times N_q) \times D_q}\). This sequence is then processed by the Video Q-Former from Video-LLaMA \cite{zhang2023video}, initialised with weights pre-trained on the HowTo-AD dataset from Movie-Llama2 \cite{han2024autoad3}. Let \(Q_V\) be a new set of randomly initialised query tokens:
\begin{equation}
    Q^{(t)}_{\text{video}} = \text{Video Q-Former}(Q_V, \tilde{Q}^{(t)}) \in \mathbb{R}^{N_q \times D_q}
    \label{eq:5}
\end{equation}

\smallskip
\noindent \textbf{Language Model Projection.} A linear projection layer projects the video embeddings \(Q^{(t)}_{\text{video}} \in \mathbb{R}^{B \times N_q \times D_q}\) to match the context window dimensions \(D_L\) of LLaMA2 \cite{touvron2023llama}:
\begin{equation}
    F = W \times Q^{(t)}_{\text{video}}, \text{ where } W \in \mathbb{R}^{D_L \times D_q}
    \label{eq:6}
\end{equation}
such that \(F \in \mathbb{R}^{B \times N_q \times D_L}\). We initialise this projection layer with the pre-trained weights from Movie-Llama2 \cite{han2024autoad3}.

\subsection{Scene-Level Visual Branch}

To improve video scene understanding, we extract global sequence representations using a memory-efficient side network-based transformer model, Side4Video (S4V) \cite{yao2023side4video}, as our feature extraction backbone. Specifically, we adopt the action recognition module from Side4Video, which operates a lightweight side network pre-trained on Kinetics-400 \cite{kay2017kinetics} in parallel with the CLIP-ViT-B/16 \cite{radford2021learning} vision model.

\smallskip
We take the subsampled video frames from \cref{eq:2} in \cref{subsec:frame-level-visual-branch} and feed each frame \(I_t\) to the Side4Video model. The parallel vision Transformer model \(g_{\text{vision}}\) splits each frame into a sequence of non-overlapping patches \(P\), which are then projected into the S4V embedding space \(D_S\). The S4V block, integrated between each Transformer layer, includes temporal convolution, [CLS] token shift \cite{yao2023side4video}, self-attention, and a fully connected MLP layer:
\begin{equation}
    S_{\text{out}} = \text{S4V}(g_{\text{vision}}(P))
\end{equation}

The resulting output from the side network \(S_\text{out} \in \mathbb{R}^{T \times P \times D_S}\) captures rich spatial-temporal information for each frame within the video sequence. The final scene-level representation is obtained by applying Global Average Pooling (GAP) \cite{yao2023side4video} over the embedded sequence:
\begin{equation}
    S = \frac{1}{T \times P} \sum_{t, \, p}S_\text{out}
    \label{eq:7}
\end{equation}
Lastly, the scene-level embeddings \(S\) are projected to the LLaMA2 embedding dimension such that \(S \in \mathbb{R}^{B \times 1 \times D_L}\).

\subsection{Dual-Vision Transformer}
\begin{figure*}[t]
  \centering
  \includegraphics[width=\linewidth]{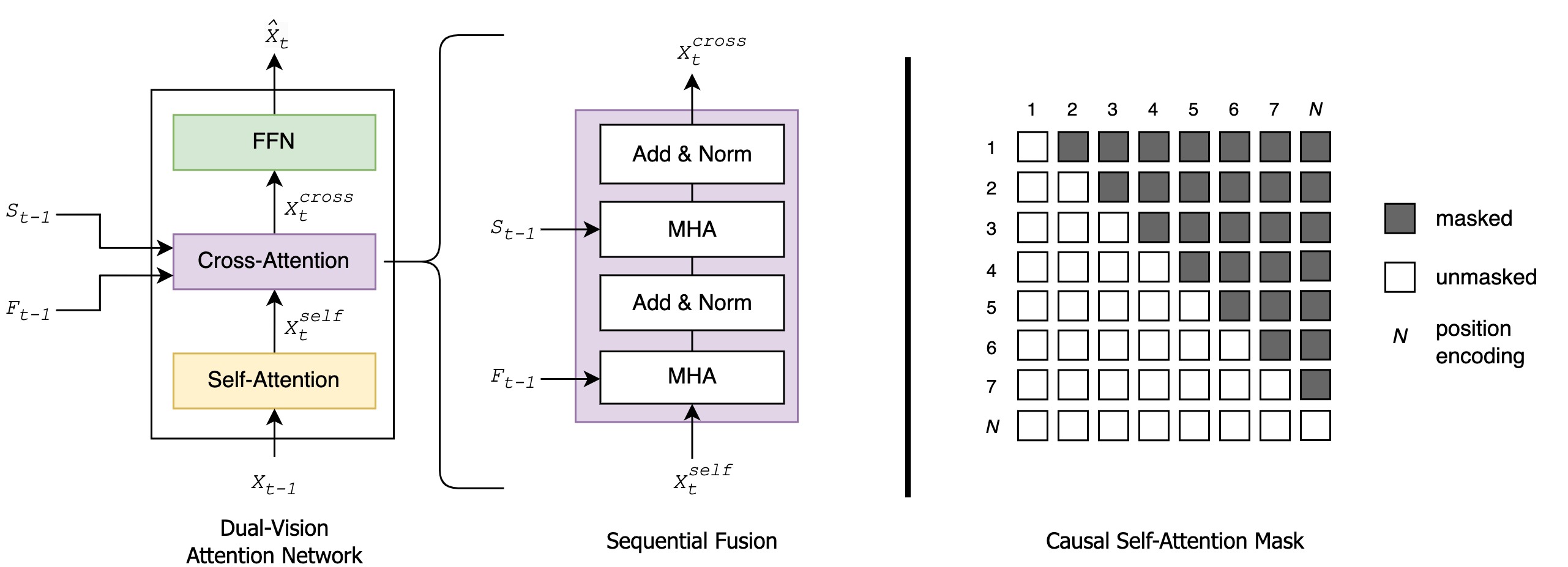}
   \caption{We propose a sequential fusion method within the Dual-Vision Attention Network to integrate frame- and scene-level embeddings. Ground-truth word embeddings are processed using a causal self-attention mask.}
   \label{fig:cross-attention}
\end{figure*}

The dual-vision Transformer network is our novel approach to integrating long-term AD context within individual video frames by cross-attending to frame and scene-level embeddings via a sequential fusion strategy \cite{nguyen2022grit}. The dual-vision network takes as input the video frame embeddings $F$ from \cref{eq:6}, global scene embeddings $S$ from \cref{eq:7}, and ground-truth AD segments. During training, the network leverages the ground-truth AD segments as supervision to enhance alignment between video embeddings and text descriptions. Specifically, the model is trained to predict the next token in the AD sequence conditioned on the video embeddings and the preceding ground-truth tokens, ensuring accurate temporal and semantic alignment. The Transformer architecture, as shown in \cref{fig:cross-attention}, consists of three layers, each featuring a self-attention module, a cross-attention module, and a Feed-Forward Network (FFN).

\smallskip
The ground-truth captions are first tokenized and encoded into \(D_L\)-dimensional word embeddings using the pre-trained LLaMA2-7B model. The Transformer generates an embedded AD sequence in an autoregressive manner, conditioning on both the visual embeddings (frame and scene-level) and the preceding word embeddings to predict the next word in the sequence. At each time step \(t\), it uses the sequence of words predicted up to time step \(t-1\) to generate the next token.

To preserve word order, sinusoidal positional embeddings~\cite{Vaswani2017AttentionisAll} are added to the word embeddings before self-attention is applied to create the position-encoded representations \(\omega\) where \(\omega \in \mathbb{R}^{D_L}\) . A causal attention mask is then used to restrict interactions with future tokens, ensuring proper autoregressive training.

\smallskip
\textbf{Sequential Cross-Attention Fusion.} Cross-attention is computed independently for each type of visual embedding using two stacked multi-head attention layers. This sequential approach allows the model to extract frame-level actions and events before integrating them into a broader scene-level narrative. Specifically, the first attention layer processes frame-level embeddings $F_{t-1}$ to capture fine-grained temporal details. The second layer then refines this representation by attending to the scene-level embeddings $S_{t-1}$, incorporating higher-level contextual cues before generating the final description.  

\smallskip
Within the multi-head attention mechanism, the frame or scene-level visual embeddings act as keys and values, while the word embeddings $\omega$ serve as queries. We exclude the ground-truth text during inference and initiate autoregressive generation with an embedded \texttt{[BOS]} token. The model then recursively feeds its own predictions back into the decoder at each timestep until the sequence is complete.

\subsection{Language Model}

The final component of our audio description generation pipeline is the natural language decoding stage. For this, we leverage the open-source LLaMA2-7B model, a pre-trained language model that remains frozen during training to preserve its generalisation capabilities and reduce computational overhead. The decoder processes the output embeddings from the dual-vision attention module, projecting them into the language model's embedding space before feeding them into LLaMA2-7B. To facilitate autoregressive text generation, we prepend a \texttt{[BOS]} embedding to indicate the start of the sequence and append a \texttt{[EOS]} embedding to signal termination, ensuring coherent and bounded outputs.

\subsection{Training Details}

Our initial film sequences \(M\) are subsampled to \(T=8\) frames, aligning with the 8-frame subsampling standard used in prior works \cite{han2023autoad2, han2024autoad3}. This consistency enables a fair and accurate comparison of our results with these baselines. The Q-Former and Video Q-Former are initialised with 32 queries, following the configurations established in \cite{han2024autoad3, zhang2023video}. During training, we only train the parameters of the frame-level projection layer, the scene-level projection layer, and the dual-vision attention network, while all other components, including the Q-Former and Video Q-Former are initialised with pre-trained weights and kept frozen. This selective training approach maintains robustness while adapting the model to our specific task.

\smallskip
Given that the model's AD generation task is extensively pre-trained on the HowTo-AD dataset \cite{han2024autoad3}, we conduct our context-aware fine-tuning for only 2 epochs to adapt the model effectively without overfitting to our smaller fine-tuning dataset. We use the AdamW optimizer \cite{loshchilov2017decoupled}, a learning rate of \(3 \times 10^{-5}\), and a cosine decay schedule to ensure stable convergence. We pre-compute the frame- and scene-level visual embeddings offline to enhance computational efficiency and load them offline during training. This offline strategy significantly reduces memory demands and computational overhead, enabling our training pipeline to fit on a single RTX-4090 GPU with 24GB of memory.

\begin{table*}[htbp]
  \centering
  \renewcommand{\arraystretch}{1.2}
  \begin{tabular}{l|cc|cc|cc}  
    \toprule
    \multirow{2}{*}{Method} & \multirow{2}{*}{VLM} & \multirow{2}{*}{LLM} & \multirow{2}{*}{Cr $\uparrow$} & \multirow{2}{*}{R@1/5$\uparrow$} & \multicolumn{2}{c}{LLM-AD-Eval (\%)$\uparrow$} \\
     & & & & & LLaMA & GPT-3.5 \\
    \midrule
    \midrule
    Video-BLIP2 \cite{yu2023videoblip} & EVA-CLIP & OPT-2.7B & 4.8 & 22.0 & 31.50 & 23.33 \\
    Video-Llama2 \cite{cheng2024videollama} & EVA-CLIP & LLaMA2-7B & 5.2 & 23.6 & 31.83 & 23.83 \\
    \midrule
    AutoAD-II \cite{han2023autoad2} & CLIP-B32 & GPT-2 & 13.5 & 26.1 & 34.66 & 25.50 \\
    AutoAD-III \cite{han2024autoad3} & EVA-CLIP & OPT-2.7B & 22.3 & 29.8 & 46.33 & 37.50 \\
    AutoAD-III \cite{han2024autoad3} & EVA-CLIP & LLaMA2-7B & 25.0 & 31.2 & 48.67 & \textbf{38.17}  \\
    DistinctAD \cite{fang2024distinctad} & CLIP\textsubscript{AD}-B16 & LLaMA3-8B & 22.7 & \textbf{33.0} & 48.00 & - \\
    \midrule
    \midrule
    \textbf{DANTE-AD} & EVA-CLIP + S4V & LLaMA2-7B & \textbf{28.89} & 28.01 & \textbf{48.83} & 34.50 \\
    \bottomrule
  \end{tabular}
  \caption{Comparisons of AD performance on the CMD-AD dataset. LLM-AD-Eval \cite{han2024autoad3} is evaluated with LLaMA2-7B-chat (left) and GPT-3.5-turbo (right). We report the results for our method DANTE-AD using the sequential fusion of our visual embeddings.}
  \label{tab:cmd-ad-all-results}
\end{table*}

\section{Experiments}
\subsection{Datasets}
We train and evaluate our method using the CMD-AD dataset \cite{han2024autoad3}, a version of the Condensed Movie Dataset (CMD) \cite{bain2020condensed} adapted explicitly for audio description. CMD-AD consists of short, key scenes (approximately 2 minutes each) from well-known films sourced from YouTube. Each scene is annotated with human-generated audio descriptions, which have been transcribed from AudioVault\footnote{https://audiovault.net} using WhisperX \cite{bain2023whisperx}. These detailed, natural language descriptions of the visual scenes serve as the ground truth for training and evaluating our model. Due to various encoding issues with the raw videos, our processed version of the CMD-AD dataset is reduced from approximately 101k down to 96k AD segments. We will be releasing the audio visual feature embeddings to allow further research on this dataset. \cref{tab:dataset} details the number of AD segments, scenes, and film counts in each split of our processed dataset compared to the original CMD-AD dataset.

\begin{table}[htbp]
  \centering
  \renewcommand{\arraystretch}{1.1}
  \begin{tabular}{cp{1.9cm}|ccr}
    \toprule
    & Version & Train & Eval & Total\\
    \midrule
    Segments & 
    CMD-AD & 93,952 & 7,316 & 101,268\\
    Segments & 
    DANTE-AD & 89,798 & 7,075 & 96,873 \\
    \midrule
    Scenes & 
    CMD-AD & 8,324 & 591 & 8,915 \\
    Scenes & 
    DANTE-AD & 8,017 & 575 & 8,592 \\
    \midrule
    Films & 
    CMD-AD & 1,321 & 98 & 1,419 \\
    Films & 
    DANTE-AD & 1,319 & 97 & 1,416 \\
    \bottomrule
  \end{tabular}
  \caption{Statistics of AD data used in our work compared to the original CMD-AD \cite{han2024autoad3} dataset.}
  \label{tab:dataset}
\end{table}

\subsection{Evaluation Metrics}

We evaluate DANTE-AD using both traditional NLP metrics (CIDEr \cite{vedantam2015cider} and Recall@k/N \cite{han2023autoad2}) and the LLM-based metric LLM-AD-Eval \cite{han2024autoad3}. CIDEr measures word-level precision between a generated sentence and a consensus of reference descriptions. It benefits tasks with high output diversity, where multiple valid descriptions can accurately depict the same visual content. In contrast, R@k/N is calculated using BertScore \cite{zhang2019bertscore}, which evaluates recall by assessing how many of the \(N\) ground-truth segments appear among the model's top-\(k\) predictions. Although these metrics are widely used in natural language evaluation, they have limitations when applied to long video descriptions, where varied phrasing for the same idea increases. We incorporate LLM-based metrics to address this, better accommodating semantically equivalent yet syntactically distinct sentences. We utilise the open-source LLaMA2-7B-chat model alongside the GPT-3.5-turbo APIs from OpenAI for LLM-AD-Eval. Each model is prompted to score the similarity between predicted audio descriptions and their ground-truth counterparts on a 0-5 scale, with 5 representing the highest similarity. These scores are then converted to percentages for clearer reporting. While LLM-AD-Eval better aligns with human judgment, the subjective interpretation of its 0-5 scores introduce ambiguity.

\subsection{Quantitative Results}
\label{subsec:quantitative-results}

We evaluate our model’s performance against prior AD methods, presenting our results in \cref{tab:cmd-ad-all-results}. For additional context, we include performance on two non-AD video captioning benchmarks: Video-BLIP2 \cite{yu2023videoblip} and Video-LLaMA2 \cite{cheng2024videollama}. Our DANTE-AD model is trained and evaluated on the CMD-AD dataset, with the model weights pre-trained on HowTo-AD \cite{han2024autoad3}. DANTE-AD outperforms all previous methods on CIDEr and LLM-AD-Eval when evaluated using LLaMA2-7B, demonstrating performance on both word level precision and relative similarity measures. 

\begin{figure}[h]
  \centering
  \includegraphics[width=\linewidth]{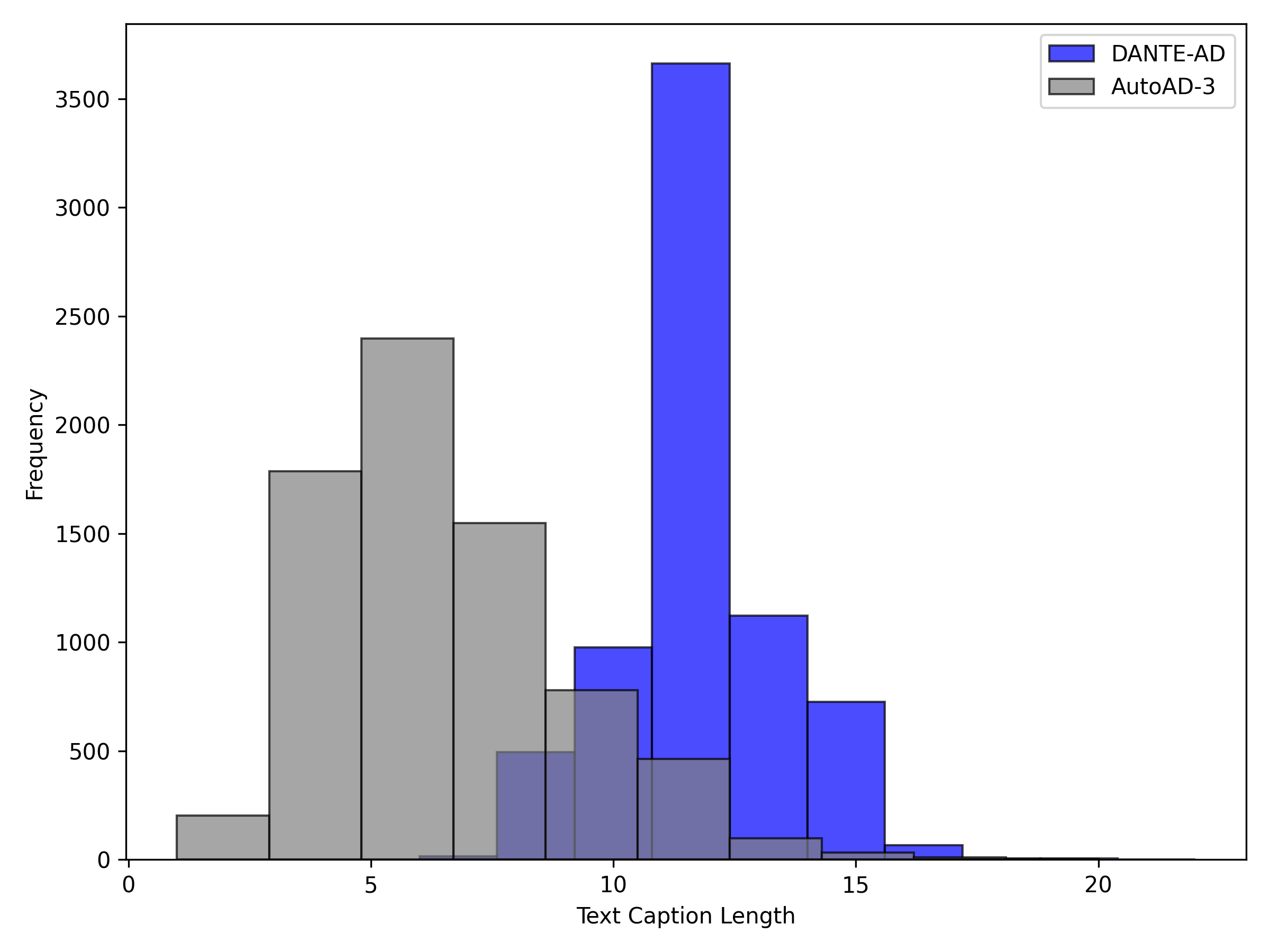}
  \caption{Text Caption length distribution of our generated descriptions (blue) compared to AutoAD-III \cite{han2024autoad3} (grey).}
  \label{fig:histogram}
\end{figure}

\begin{figure*}[t]
  \centering
  \includegraphics[width=\linewidth]{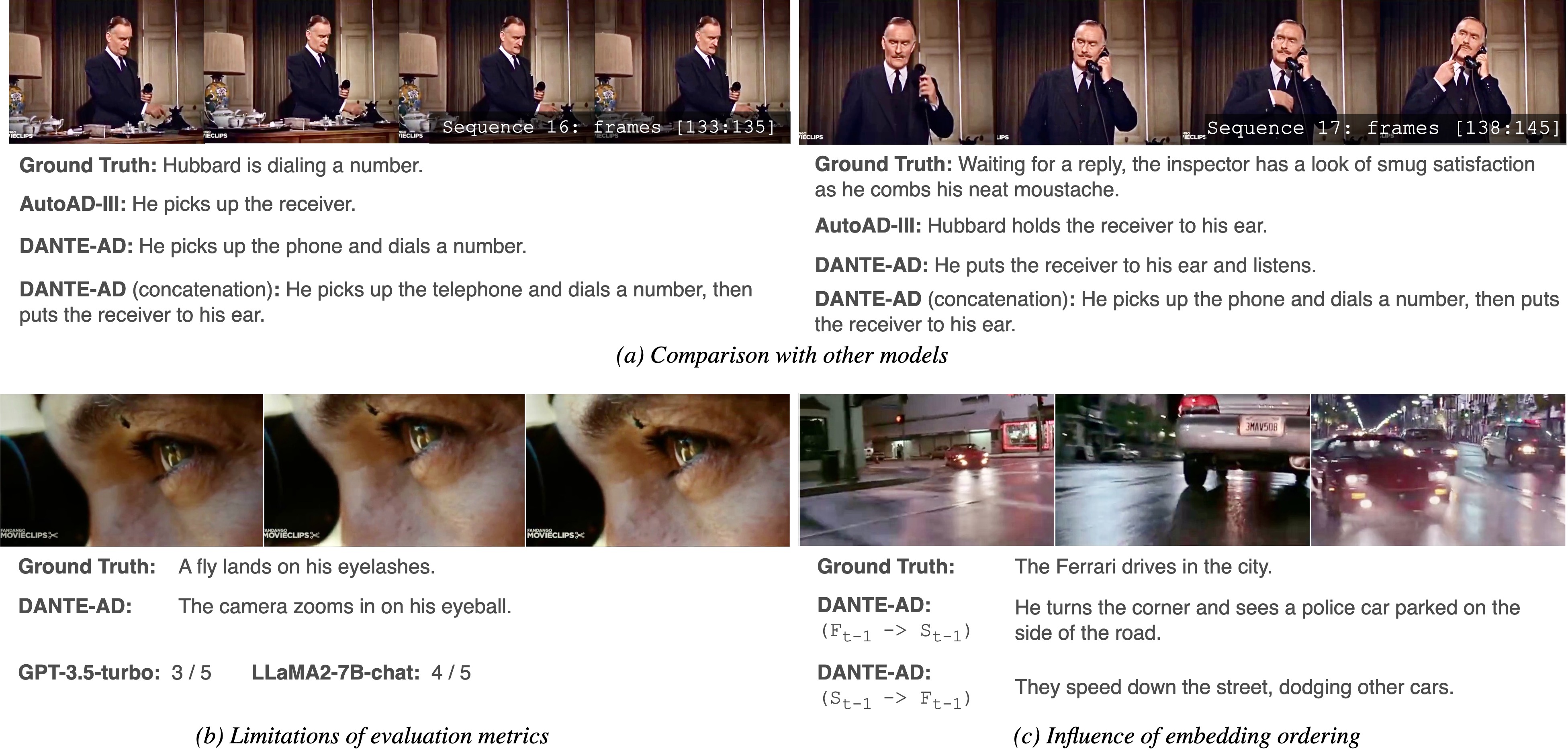}
  \caption{\textbf{Qualitative results on CMD-AD-Eval.} (a) Two consecutive AD segments. (``Dial M for Murder: 2017/JY4UoItJ\_lA"), compared with the naive concatenation method and AutoAD-III. (b) Comparison of the qualitative results with quantitative LLM-AD-Eval scores on the two language models (``The Hurt Locker: 2015/rviQWy48B\_w"). (c) Qualitative results comparing the influence of frame- and scene-level embedding ordering within the sequential fusion cross-attention (``Bulletproof: 2011/\_d4H6lx9-Is").}
  \label{fig:qualitative}
\end{figure*}

\smallskip
\noindent \textbf{Effect of text caption length.} For AD, a more extended caption is often desired to provide storytelling elements. Therefore, we analyse the caption length on DANTE-AD compared to AutoAD-III. As shown in \cref{fig:histogram}, our method generates descriptions that average longer than those produced by AutoAD-III. We infer that the additional scene-level context enhances the generated descriptions, enabling greater detail due to the richer information provided. This is further supported by our qualitative results in \cref{fig:qualitative}, which compares a shorter sentence from AutoAD-III to our method’s more fine-grained output.

\begin{figure}[htbp]
  \centering
  \includegraphics[width=\linewidth]{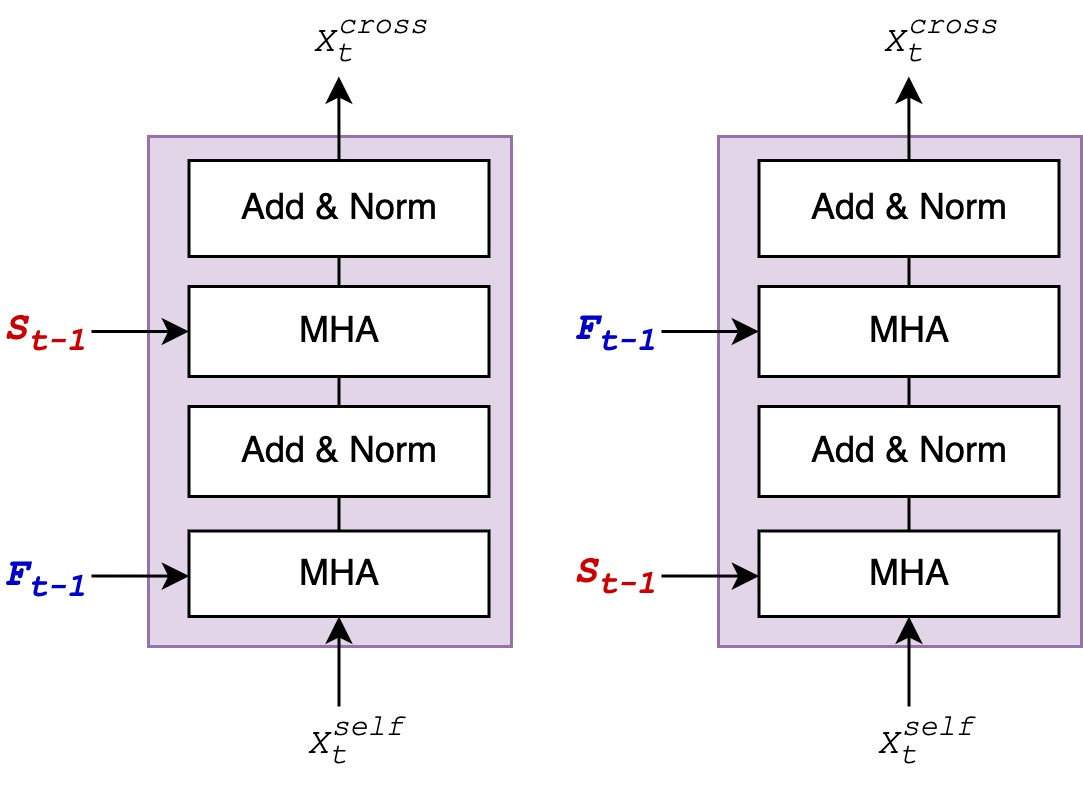}
  \caption{Illustration of the ordering of frame-level (\(F_{t-1}\)) and scene-level (\(S_{t-1}\)) visual embeddings during sequential cross-attention within our dual-vision attention network. }
  \label{fig:figure-ablation}
\end{figure}

\smallskip
\noindent \textbf{Effect of visual embedding ordering.} The sequential fusion of our method is key to integrating the scene and frame details. Therefore, we can explore various configurations of the cross-attention to assess the causal relationship between scene- and frame-level processing, as illustrated in \cref{fig:figure-ablation}. Specifically, we evaluate whether processing frame-level embeddings first (\(F_{t-1} \implies S_{t-1}\)) improves scene comprehension or if prioritising scene-level context (\(S_{t-1} \implies F_{t-1}\)) improves contextual grounding for frame-level representations. Previous findings in \cref{tab:cmd-ad-all-results} indicate that the LLM-AD-Eval metric yields higher scores for lengthy, verbose descriptions, regardless of their fidelity to the video content. In contrast, the CIDEr metric, which relies on word-for-word alignment with ground-truth descriptions, prioritises accuracy relative to the video content. Consequently, \cref{tab:sequential-xatt-ablation} reveals that initialising frame-level information with global context first improves contextual precision. Conversely, generating global information after frame-level details produces more elaborate, detailed sentences, as qualitatively demonstrated in \cref{fig:qualitative}(c).

\begin{table}[htbp]
    \centering
    \renewcommand{\arraystretch}{1}
    \resizebox{\linewidth}{!}{
        \begin{tabular}{c|c|c|cc}
            \toprule
            \multirow{2}{*}{Method} & \multirow{2}{*}{Cr $\uparrow$} & \multirow{2}{*}{R@ $\uparrow$} & \multicolumn{2}{c}{LLM-AD-Eval (\%) $\uparrow$}   \\
                &    &   & LL      &G3.5                        \\
            \midrule
            \(S \implies F\) &\textbf{35.71} & 27.78 & 44.17 & 33.33 \\
           \rowcolor{gray!20!} \(F \implies S\) & 28.89 & \textbf{28.01} & \textbf{48.83} & \textbf{34.50} \\
            \bottomrule
        \end{tabular}
    }
    \caption{Results of sequential cross-attention in our dual-vision model, comparing the impact of the order of frame- (\(F\)) and scene-level (\(S\)) embeddings on AD performance. CIDEr and LLM-AD-Eval reported using LLaMA2-7B and GPT-3.5.}
    \label{tab:sequential-xatt-ablation}
\end{table}

\begin{figure}[htbp]
  \centering
    \centering
    \includegraphics[width=\linewidth]{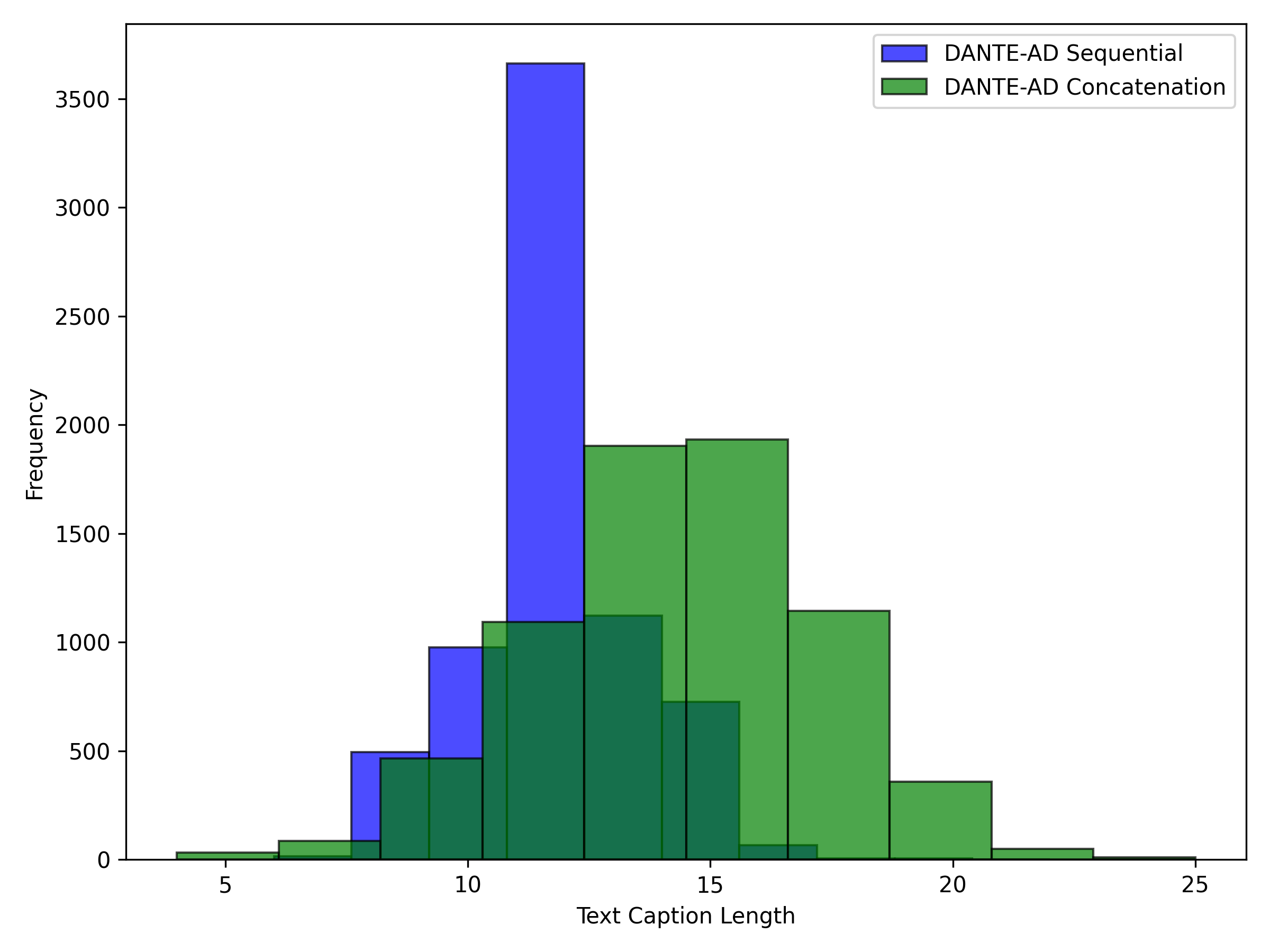}
    \caption{Comparison of the text caption length distribution of generated descriptions between DANTE-AD using concatenation and the sequential fusion method.}
    \label{subfig:figure-ablation-hist}
\end{figure}

\smallskip
\noindent \textbf{Concatenated cross-attention.} Alongside the sequential fusion of the scene and frame embeddings, we compare to simply using the concatenation of the embeddings along their video sequence length, forming the fused visual embeddings \([F; S]\). These are the keys and values in the multi-head attention layer, while the position-encoded word embeddings \(\omega\) serve as the queries. \cref{subfig:figure-ablation-hist} shows the distribution of the caption lengths for concatenation and our proposed approach. The naive concatenation method generates the longest text captions, as shown in \cref{subfig:figure-ablation-hist}. However, these extended captions often go beyond the video's visual content, introducing verbose captions that do not align with the actual temporal localisation. For example, in sequence 16 in \cref{fig:qualitative}(a), the character is only dialling the phone, and it is not until sequence 17 that he puts the receiver to his ear. In contrast, our sequential method yields the most accurate results. Segment 16 provides a more descriptive caption than the ground truth, while sequence 17 focuses on the beginning of the scene rather than the latter part where the moustache combing is shown. This highlights a limitation of evaluation metrics that rely solely on comparisons with the ground-truth text, as the visual information being described is correct, albeit presented from a different perspective. This is quantitatively shown with the concatenation achieving a CIDEr score of only 21, compared to our 28.9. 

\subsection{Qualitative Results}

Alongside our quantitative comparison, \cref{fig:qualitative} compares the ground truth, the simple concatenation approach, and the outputs of AutoAD-III \cite{han2024autoad3}. This demonstrates that our approach produces descriptions that are neither too brief or verbose. \cref{fig:qualitative}(b) highlights the challenge of using the LLM-based similarity metrics, as it results in a wide range of scores despite our output qualitatively aligning well with the ground truth. Similarly, \cref{fig:qualitative}(c) underscores the importance of ordering in sequential fusion, our method, $(F \implies S)$, generating more reasoned and descriptive audio descriptions. Further examples of generated audio description captions are provided in the supplementary material in section \cref{sec:Suppl-qualitative}.

\section{Conclusion}
\label{sec:conclusion}

In this work, we introduced \textbf{DANTE-AD}, a novel dual-vision Transformer-based model for generating detailed and contextually rich audio descriptions in long-form video content. Unlike prior methods that rely solely on frame-level embeddings, DANTE-AD integrates both frame- and scene-level representations through a sequential fusion strategy, enabling a more comprehensive understanding of temporal context. By incorporating a multi-stage attention mechanism, our model effectively grounds fine-grained visual details within a coherent long-term narrative.

We extensively evaluated real-world movie clips and demonstrated that DANTE-AD significantly outperforms existing audio description techniques, providing more emotive and story-rich results. Our findings highlight the importance of leveraging spatial and temporal modelling capabilities to bridge the gap between object-centric video captioning and holistic storytelling in AD generation.

Future work may explore further refinements in long-term scene understanding, adaptive attention mechanisms for balancing local and global information, and enhanced multimodal fusion techniques to incorporate additional modalities, such as audio and subtitles. We hope our contributions pave the way for more effective and accessible automated AD solutions, improving media accessibility for vision-impaired audiences.

\bibliographystyle{ieeenat_fullname}
\bibliography{main}

\clearpage
\setcounter{page}{1}
\maketitlesupplementary

\vspace{-0.5cm}
\section{LLM-AD-Eval Prompts}
\label{sec:llm_eval_prompts}

For accurate comparison with previous methods, we use the same prompts for LLM-AD-Eval as provided in \cite{han2024autoad3}. For reference, the prompt for LLM-AD-Eval using LLaMA2-7B-chat is given in \cref{fig:llama-prompt} and for GPT-3.5-turbo in \cref{fig:gpt-prompt}.

\begin{figure}[htbp]
  \centering
  \fbox{
  \renewcommand{\arraystretch}{1.1}
  \begin{tabular}{p{0.95\linewidth}}
    Please evaluate the following movie audio description pair: \\
    - Correct Audio Description: \{\textit{ground-truth AD segment}\} \\
    - Predicted Audio Description: \{\textit{predicted AD segment}\} \\
    Provide your evaluation only as a matching score where the matching score is an integer value between 0 and 5, with 5 indicating the highest level of match. \\
    Please generate the response in the form of a Python dictionary string with keys 'score', where its value is the matching score in INTEGER, not STRING. \\
    DO NOT PROVIDE ANY OTHER OUTPUT TEXT OR EXPLANATION. Only provide the Python dictionary string. For example, your response should look like this: \{`score': \}. \\
  \end{tabular}
  }
  \caption{LLM-AD-Eval prompt for evaluations using LLaMA2-7B-chat from \cite{han2024autoad3}.}
  \label{fig:llama-prompt}
\end{figure}

\begin{figure}[htb!]
  \centering
  \renewcommand{\arraystretch}{1.1}
  \fbox{
  \begin{tabular}{p{0.95\linewidth}}
    System: \\
    You are an intelligent chatbot designed for evaluating the quality of generative outputs for movie audio descriptions. Your task is to compare the predicted audio descriptions and determine its level of match, considering mainly the visual elements like actions, objects and interactions. Here's how you can accomplish the task: \\
    Instructions: \\
    - Check if the predicted audio description covers the main visual elements from the movie, especially focusing in the verbs and nouns. \\
    - Evaluate whether the predicted audio description includes specific details rather than just generic points. It should provide comprehensive information that is tied to specific elements of the video. \\
    - Consider synonyms or paraphrases as valid matches. Consider pronouns like `he' or `she' as valid matches with character names. Consider different character names as valid matches. \\ 
    - Provide a single evaluation score that reflects the level of match of the prediction, considering the visual elements like actions, objects and interactions.
    
    \medskip
    User: \\
    Please evaluate the following movie description pair: \\
    - Correct Audio Description: \{\textit{ground-truth AD segment}\} \\
    - Predicted Audio Description: \{\textit{predicted AD segment}\} \\
    Provide your evaluation only as a matching score where the matching score is an integer value between 0 and 5, with 5 indicating the highest level of match.
    Please generate the response in the form of a Python dictionary string with keys `score', where its value is the matching score in INTEGER, not STRING. \\
    DO NOT PROVIDE ANY OTHER OUTPUT TEXT OR EXPLANATION. Only provide the Python dictionary string. For example, your response should look like this: \{`score': \}. \\
  \end{tabular}
  }
  \caption{LLM-AD-Eval prompt for evaluations using GPT-3.5-turbo from \cite{han2024autoad3}.}
  \label{fig:gpt-prompt}
\end{figure}

\section{Qualitative Examples}
\label{sec:Suppl-qualitative}

\begin{figure*}[h]
  \centering
    \centering
    \includegraphics[width=0.95\linewidth]{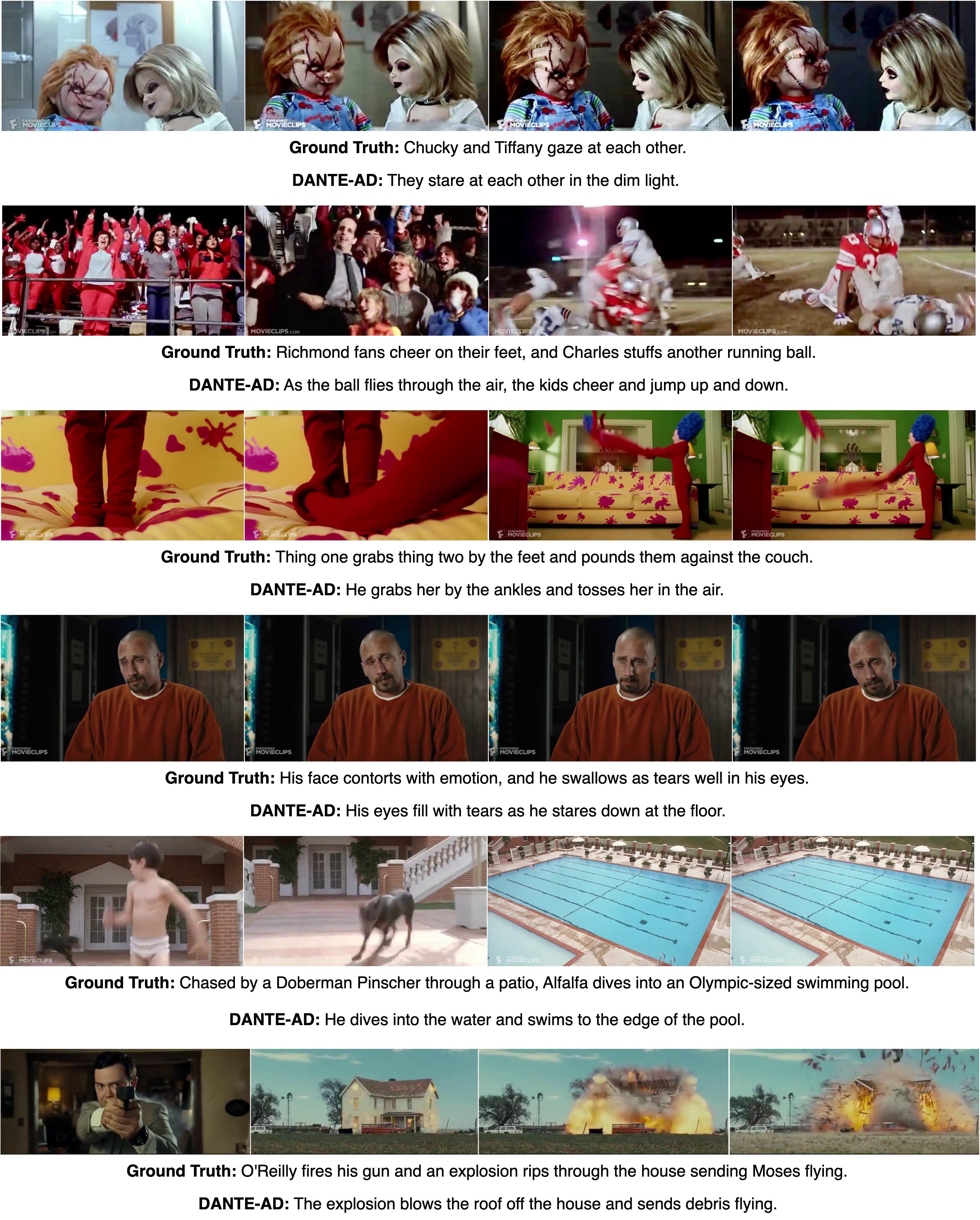}
    \caption{Qualitative results of our DANTE-AD method on CMD-AD-Eval. Our method uses sequential fusion cross-attention between frame- and scene-level visual embeddings.}
    \label{fig:supp-qualitative}
\end{figure*}

To showcase our results on DANTE-AD, we provide additional qualitative examples in \cref{fig:supp-qualitative}.

\end{document}